\documentclass[conference]{IEEEtran}
\IEEEoverridecommandlockouts

\usepackage[square,numbers, sort&compress]{natbib}
\usepackage[utf8]{inputenc} 
\usepackage[T1]{fontenc}    
\usepackage[draft]{hyperref}       
\usepackage{url}            
\usepackage{booktabs}       
\usepackage{nicefrac}       
\usepackage{xcolor}         
\usepackage{graphicx}
\usepackage{mathtools}
\usepackage{scalerel,amssymb}
\usepackage{subfig}
\usepackage{comment}
\usepackage{amsmath,amssymb,amsfonts}
\usepackage{algorithmic}
\usepackage{textcomp}
\def\msquare{\mathord{\scalerel*{\Box}{gX}}}
\def\BibTeX{{\rm B\kern-.05em{\sc i\kern-.025em b}\kern-.08em
		T\kern-.1667em\lower.7ex\hbox{E}\kern-.125emX}}

\begin{document}
\title{Addressing the Inconsistency in Bayesian Deep Learning via Generalized Laplace Approximation}
\author{\IEEEauthorblockN{1\textsuperscript{st} Yinsong Chen}
	\IEEEauthorblockA{\textit{School of Engineering} \\
		\textit{Deakin University}\\
		Melbourne, VIC, 3216, Australia \\
		yinsong.chen@deakin.edu.au}
	\and
	\IEEEauthorblockN{2\textsuperscript{nd} Samson S. Yu}
	\IEEEauthorblockA{\textit{School of Engineering} \\
		\textit{Deakin University}\\
		Melbourne, VIC, 3216, Australia \\
		s.yu@ieee.org}
	\and
	\IEEEauthorblockN{3\textsuperscript{rd} Zhong Li}
	\IEEEauthorblockA{\textit{Faculty of Mathematics and Computer Science} \\
		\textit{FernUniversität in Hagen}\\
		Hagen, 58084, Germany \\
		zhong.li@fernuni-hagen.de}
	\and
	\IEEEauthorblockN{4\textsuperscript{th} Chee Peng Lim}
	\IEEEauthorblockA{\textit{Department of Computing Technologies} \\
		\textit{Swinburne University of Technology}\\
		Hawthorn, VIC, 3122, Australia \\
		cplim@swin.edu.au}
}

\maketitle

\begin{abstract}
In recent years, inconsistency in Bayesian deep learning has attracted significant attention. Tempered or generalized posterior distributions are frequently employed as direct and effective solutions. Nonetheless, the underlying mechanisms and the effectiveness of generalized posteriors remain active research topics. In this work, we interpret posterior tempering as a correction for model misspecification via adjustments to the joint probability, and as a recalibration of priors by reducing aleatoric uncertainty. We also introduce the generalized Laplace approximation, which requires only a simple modification to the Hessian calculation of the regularized loss and provides a flexible and scalable framework for high-quality posterior inference. We evaluate the proposed method on state-of-the-art neural networks and real-world datasets, demonstrating that the generalized Laplace approximation enhances predictive performance.  
\end{abstract}


\section{Introduction}
\label{sec:1}
Bayesian inference stands as a pivotal scientific learning framework utilized today. Its fundamental tenet involves employing probability to quantify all forms of uncertainty within a statistical model and subsequently updating this uncertainty in light of observed data using Bayes theorem \cite{jospin2022hands},
\begin{equation}
	\label{eq:1}
	p(\mathcal{H}|\mathcal{D}) = \frac{p(\mathcal{D}|\mathcal{H})p(\mathcal{H})}{p(\mathcal{D})} = \frac{p(\mathcal{D}|\mathcal{H})p(\mathcal{H})}{\int_\mathcal{H} p(\mathcal{D}|\mathcal{H}')p(\mathcal{H}')d\mathcal{H}'}.
\end{equation} 

Here, $\mathcal{H}$ represents a hypothesis with associated prior belief, while $\mathcal{D} := \{(x_n,y_n)\}_{n=1}^{N}$ denotes $N$ observed data samples informing one's belief about $\mathcal{H}$. $p(\mathcal{H})$ represents the prior, $\int_\mathcal{H} p(\mathcal{D}|\mathcal{H}')p(\mathcal{H}')d\mathcal{H}'$ is the evidence, and $p(\mathcal{H}|\mathcal{D})$ is referred to as the posterior.   

In the past decade, it has become evident that Bayesian inference can exhibit inconsistent behavior when confronted with model misspecification \cite{grunwald2007suboptimal,erven2007catching,muller2013risk,holmes2017assigning,grunwald2017inconsistency,yao2018using,syring2019calibrating,grunwald2020fast}.For instance, reference \cite{grunwald2017inconsistency} demonstrated the inconsistency of Bayesian inference when models are misspecified, exemplified by applying a standard linear model assuming homoskedasticity to heteroskedastic data without outliers. As sample size increases, the posterior probability is observed to progressively support more complex yet inferior models.

A prevalent solution to address model misspecification involves augmenting Bayesian inference with a learning rate parameter denoted as $\mathcal{T}$, also referred to as temperature in Gibbs posterior literature:
\begin{equation}
	\label{eq:2}
	p(\mathcal{H}|\mathcal{D}) = \frac{p(\mathcal{D}|\mathcal{H})^{\mathcal{T}}p(\mathcal{H})}{\int_\mathcal{H} p(\mathcal{D}|\mathcal{H}')^{\mathcal{T}}p(\mathcal{H}')d\mathcal{H}'}.
\end{equation}

This method, termed generalized or tempered Bayes, has been proposed independently by multiple authors \cite{barron1991minimum,walker2001bayesian,zhang2006varepsilon,grunwald2012safe,bissiri2016general}. The role of $\mathcal{T}$ is evident; for $0<\mathcal{T}<1$, the impact of the prior is emphasized more than in the standard Bayesian update, resulting in reduced influence from the data. In contrast, for $\mathcal{T}>1$, the likelihood gains prominence over the standard Bayesian update, and in extreme cases with very large $\mathcal{T}$, the posterior concentrates on the maximum likelihood estimator for the model. In the extreme case when $\mathcal{T}=0$, the posterior consistently equals the prior. In statistical Bayesian inference, $\mathcal{T}$ is typically selected to be less than one \cite{heide2020safe}. 
 
A comparable challenge and resolution have been identified in the realm of Bayesian deep learning \cite{guo2017calibration,wenzel2020good, osawa2019practical}. For instance, Wenzel et al. \cite{wenzel2020good} reported that Bayesian model averaging exhibits inferior performance compared to standard deep learning, a phenomenon known as the cold posterior effect. They demonstrated that introducing a generalized posterior of the form
\begin{equation}
	\label{eq:3}
	p(\theta|\mathcal{D}) = \frac{p(\mathcal{D}|\theta)^{\mathcal{T}}p(\theta)^{\mathcal{T}}}{\int_\theta p(\mathcal{D}|\theta')^{\mathcal{T}}p(\theta')^{\mathcal{T}}d\theta'},
\end{equation} can improve predictive performance, where $\theta$ is the model parameterization, typically representing the hypothesis $\mathcal{H}$ in Bayesian deep learning. Assuming Gaussian priors over the model parameters, i.e., $p(\theta) = \mathcal{N}(0,\sigma)$, as is often the case in Bayesian neural networks, Eq. (\ref{eq:3}) is essentially Eq. (\ref{eq:2}) with rescaled prior variances \cite{aitchison2020statistical}. The derivation is straightforward:
\begin{equation}
\label{eq:4}
\begin{split}
	\frac{1}{\mathcal{T}}\log p(\theta) & = -\frac{1}{2\mathcal{T}\sigma^2} \sum_{n=1}^{N} \theta_n^2 + const. \\
	& = -\frac{1}{2(\sqrt{\mathcal{T}}\sigma)^2} \sum_{n=1}^{N} \theta_n^2 + const. = \log p'(\theta), \\
	p'(\theta) & = \mathcal{N}(0,\sqrt{\mathcal{T}}\sigma).
\end{split}
\end{equation} 

In this study, we differentiate Bayesian deep learning from conventional statistical Bayesian inference. While they share the same theoretical foundation, Bayesian deep learning typically operates with larger datasets and more complex data structures, necessitating scalable learning methodologies such as the Hamiltonian Monte Carlo algorithm \cite{neal2011mcmc}, variational inference (VI) \cite{blei2017variational}, Laplace approximation (LA) \cite{daxberger2021laplace}, and sophisticated models like deep neural networks. In contrast to identifying model misspecification as the cause of inconsistency in Bayesian inference, Bayesian deep learning identifies inadequate prior selection as a potential source of inconsistency \cite{wenzel2020good, aitchison2020statistical, chen5100162interplay}. Furthermore, as documented in the Bayesian deep learning literature \cite{zhang2018noisy, leimkuhler2019partitioned, heek2019bayesian, zhang2020cyclical}, $\mathcal{T}$ is commonly chosen to exceed one. However, while the effectiveness of generalized Bayes in addressing model misspecification in Bayesian inference has been extensively studied, investigations into its ability to resolve the cold posterior effect in Bayesian deep learning remain underexplored.

Additionally, generalized Bayesian methods are primarily implemented within Markov chain Monte Carlo and VI frameworks \cite{wenzel2020good}, whereas LA methods are generally confined to the standard Bayesian posterior and lack support for temperature scaling or generalized Bayes formulations. As a result, these methods are unable to directly mitigate the cold posterior effect, which has been shown to negatively impact the predictive accuracy and uncertainty quantification of Bayesian neural networks.

Our primary contributions are twofold: 1) a comprehensive review and extension of the theoretical framework explaining how the generalized posterior mitigates model misspecification and suboptimal priors (Section \ref{sec:2}), and 2) the development of a generalized LA (GLA) method that integrates generalized Bayes principles, along with empirical evaluation of its properties. (Section \ref{sec:3}). Our experimental results (Section \ref{sec:4}) demonstrate that the GLA can enhance performance on in-distribution data with an appropriately chosen $\mathcal{T}$, though the optimal $\mathcal{T}$ is significantly influenced by the selected Hessian approximation. Additionally, we show the susceptibility of the GLA to out-of-distribution data caused by domain shift.

\section{How Generalized Bayes Can Resolve Model Misspecification and Poor Prior}
\label{sec:2}  
\subsection{How Generalized Bayes Can Resolve Model Misspecification}
\begin{figure}
	\centering
	\includegraphics[width=0.33\textwidth]{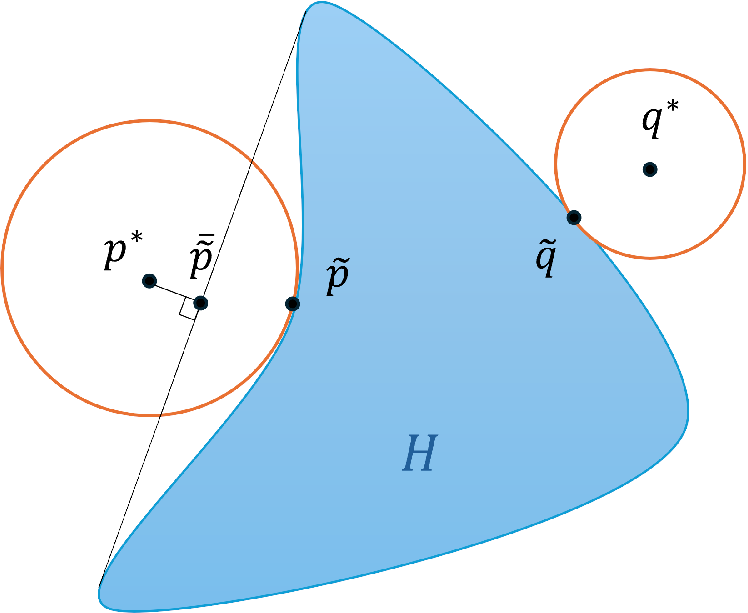}
	\caption{Model misspecification.}
	\label{fig:1}
\end{figure}

To gain a clearer understanding of model misspecification, let us commence with an intuitive example initially presented in \cite{grunwald2017inconsistency}. As depicted in Fig. \ref{fig:1}, when the 'ground truth' $p^*$ lies beyond the hypothesis space $\mathcal{H}$, indicating model misspecification, and the model is non-convex, the distribution $\bar{\tilde{p}}(y|x)$ minimizing the Kullback-Leibler (KL) divergence to the 'ground truth' $p^*$ within the convex hull of $\mathcal{H}$ can significantly differ from $\tilde{p}$. Here, $\tilde{p}$ denotes the distribution in the hypothesis space that minimizes KL divergence to the 'true' $p^*$. In cases where $q^*$ rather than $p^*$ represents the true distribution, the convexity property enables $\tilde{q}$ to attain the minimum. Therefore, one trivial explanation for why Bayesian deep learning does not experience issues with model misspecification could be that, in cases of overparameterization, the optimization landscape of neural networks is nearly convex \cite{du2018gradient,li2018learning,allen2019learning}.

Reference \cite{grunwald2017inconsistency} offers an explanation for the efficacy of the $\mathcal{T}$-generalized posterior in correcting Bayesian updates when models are misspecified: for sufficiently small $\mathcal{T}$, it is formally identical to a conventional posterior derived from an adjusted joint probability model. To delve into this interpretation, we start with specific settings. A learning problem is defined by a tuple $(l, \mathcal{P}, \mathcal{F})$ \cite{vapnik2013nature}, where $\mathcal{P}$ represents distributions over a sample space $\mathcal{Z} := \mathcal{X} \times \mathcal{Y}$, $\mathcal{F}$ denotes a set of models (also known as the hypothesis space $\mathcal{H}$ in certain Bayesian literature), and $l := \mathcal{F} \times \mathcal{Z} \rightarrow \mathbb{R} \cup \{\infty\}$ denotes a loss function. 

Let $h^*(x,y)$ denote the density function of the true distribution $p*$ on $Z:=(X,Y)\in \mathcal{P}$. The $\mathcal{T}$-generalized distributions $p^{(\mathcal{T})}$ are formally defined as joint distributions over $Z$ with density functions $h^{(\mathcal{T})}$: 
\begin{equation}
	\label{eq:5}
	h^{(\mathcal{T})}(x,y|\theta) = h^*(x,y)\dot ( \frac{h(y|x,\theta)}{h(y|x,\tilde{\theta})})^{\mathcal{T}}, 
\end{equation} extended to sample size by independence, where $\tilde{\theta}$ minimizes the loss function, typically represented by the negative log-likelihood (NLL) in our context. To derive the interpretation, we introduce an auxiliary definition.

\textbf{Definition 1 (Central Condition (\cite{grunwald2020fast}))} For $\mathcal{T}>0$, $(l, \mathcal{P}, \mathcal{F})$ satisfies the $\mathcal{T}$-strong central condition if there exists a $\tilde{f} \in \mathcal{F}$ such that 
\begin{equation}
	\label{eq:7}
	E[e^{\mathcal{T}(l(\tilde{f}, Z)-l(f,Z))}]\leq 1 \text{, for all $f \in \mathcal{F}$ and $Z \in \mathcal{P}$}. 
\end{equation}

\textbf{Lemma 1} \textit{In a learning problem adhering to the $\tilde{\mathcal{T}}$-strong central condition, selecting any $0<\mathcal{T}\leq \tilde{\mathcal{T}}$ results in the following:}
\begin{equation}
	\label{eq:6}
	\int \int h^{(\mathcal{T})}(x,y|\theta) dxdy \leq 1, \textit{ for any $\theta$ in hypothesis space}.  
\end{equation}

\textbf{Proof of Lemma 1.} Substituting $l$ with NLL function and $\mathcal{T}$ with $\tilde{\mathcal{T}}$ into Eq. (\ref{eq:7}), we obtain
\begin{equation}
	\label{eq:8}
	\begin{split}
		& E_{(x,y) \sim (X,Y)}[e^{\tilde{\mathcal{T}}(-log(h(y|x, \tilde{\theta}))+log(h(y|x, \theta)))}] \\
		& = E_{(x,y) \sim (X,Y)}[( \frac{h(y|x,\theta)}{h(y|x,\tilde{\theta})})^{\tilde{\mathcal{T}}}] \leq 1.
	\end{split}
\end{equation} Based on the condition $\int \int h^*(x,y)dxdy = 1$, Eq. (\ref{eq:6}) can be readily derived. 

When Eq. (\ref{eq:6}) equals 1, the model is considered correctly specified, rendering the use of $\mathcal{T}$ unnecessary. If Eq. (\ref{eq:6}) is strictly less than 1, denoted as $1-\epsilon$, then the corresponding $p^{(\mathcal{T})}$ can be interpreted as a standard probability distribution. This interpretation involves defining it on an extended outcome space $(\mathcal{X} \times \mathcal{Y}) \cup \{\msquare\}$ and assuming that it assigns probability mass $\epsilon$ to a pseudo outcome $\msquare$, which in reality would not occur. Thus, the $\mathcal{T}$-generalized Bayes can be understood as updating the posterior using the likelihood derived from a correctly specified probability model.

Notably, the strong central condition is trivially fulfilled for density estimation when the model is accurately specified. Additionally, it automatically holds under a convex model in cases of model misspecification \cite{van2015fast}. Furthermore, for classification and other cases involving bounded excess loss, it can be linked to the Bernstein condition \cite{bartlett2005local}. 

\subsection{How Generalized Bayes Can Resolve Poor Prior}
In contemporary deep learning, commonly used datasets are typically well-curated and thus exhibit low aleatoric uncertainty \cite{aitchison2020statistical}. As a result, a generalized posterior primarily corrects priors that inaccurately assume high aleatoric uncertainty in such data. For simplicity, we consider a classification setting, where aleatoric uncertainty, $p(y|f(x,\theta^*))$, can be interpreted as the probability of observing the same input $x$ twice with different one-hot labels $y$ and $y'$:
\begin{equation}
	p(y' \neq y | f(x,\theta^*))) = E_{\theta}[\sum_{y' \neq y} \text{softmax}(f(x,\theta))_{y'} p(\theta|x,y)],
\end{equation} where $\theta^*$ is the unknown ground-truth parameters and $f(x,\theta)$ denotes the model output for parameterization $\theta$ and input $x$. Replacing the posterior $p(\theta|x,y)$ with a generalized version $p_{\mathcal{T}}(\theta|x,y)$ from Eq. (\ref{eq:3}), the probability $p_T(y' \neq y| f(x, \theta^*))$ becomes a function of $\mathcal{T}$. It has been shown that reducing $\mathcal{T}$ consistently decreases $p_{\mathcal{T}}(y' \neq y| f(x, \theta^*))$, thereby lowering aleatoric uncertainty \cite{adlam2020cold, chen5100162interplay}. 

\section{Generalized Laplace Approximation} 
\label{sec:3}
While generalized Bayes addresses model misspecification and poor priors in theory, its practical impact on deep neural networks merits attention. In high-dimensional, over-parameterized models, standard Bayesian posteriors often yield miscalibrated uncertainties, i.e., the cold posterior effect, due to mismatched priors and data. By incorporating temperature scaling into the LA, we propose the GLA as an efficient approach to adjust posterior concentration and uncertainty.
\subsection{Laplace Approximation}
Supervised deep learning operates within the framework of empirical risk minimization. Given an independent and identically distributed (i.i.d.) dataset $\mathcal{D}:= \{(x_n,y_n)\}_{n=1}^N$ where $x_n \in \mathbb{R}^I$ represent inputs and $y_n \in \mathbb{R}^O$ denote outputs, a neural network $f(x, \theta):\mathbb{R}^I \rightarrow \mathbb{R}^O$ with parameters $\theta \in \mathbb{R}^P$ is trained to minimize empirical risk. This involves minimizing a loss function expressed as the sum of empirical loss terms $\ell(x_n,y_n,\theta)$ and a regularizer $r(\theta)$ 

\begin{equation}
	\label{eq:12}
	\begin{split}
		\theta_{\text{MAP}} & = \text{argmin}_{\theta \in \mathbb{R}^P}\mathcal{L}(\mathcal{D},\theta) \\
		& = \text{argmin}_{\theta \in \mathbb{R}^P}(\sum_{n=1}^{N}\ell(x_n,y_n,\theta)+r(\theta)).
	\end{split}
\end{equation}

From a Bayesian perspective, these components correspond to i.i.d. log-likelihoods and a log-prior, respectively \cite{chen2024sparse}, confirming that $\theta_{\text{MAP}}$ represents a maximum a posteriori (MAP) estimate:
\begin{equation}
	\label{eq:13}
	\ell(x_n,y_n,\theta) = -\log p(y_n|x_n,\theta) \text{    and    } r(\theta) = -\log p(\theta).
\end{equation}

For instance, the commonly employed weight regularizer $r(\theta)=\frac{1}{2}\beta^{-2}||\theta||^2$ (also known as L2 regularizer) aligns with a Gaussian prior $p(\theta)=\mathcal{N}(0,\beta^2 I)$, while the cross-entropy loss represents a categorical likelihood. Therefore, the exponential of the negative training loss $\text{exp}(-\mathcal{L}(\mathcal{D},\theta))$ denotes an unnormalized posterior distribution. This can be demonstrated by substituting Eq. (\ref{eq:13}) into Eq. (\ref{eq:1}) and deriving
\begin{equation}
	\label{eq:14}
	p(\theta|\mathcal{D}) = \frac{p(\mathcal{D}|\theta)p(\theta)}{\int_{\theta'} p(\mathcal{D}|\theta')p(\theta')d\theta'} = \frac{1}{Z} \text{exp}(-\mathcal{L}(\mathcal{D},\theta)),  
\end{equation} where the evidence $Z = \int_{\theta'} p(\mathcal{D}|\theta')p(\theta')d\theta'$ denotes a constant normalization factor. 

The LA \cite{mackay1991bayesian} utilizes a second-order Taylor expansion of $\mathcal{L}$ at the mode $\theta_{\text{MAP}}$, resulting in an approximation:
\begin{equation}
	\label{eq:15}
		\mathcal{L}(D,\theta) \approx \mathcal{L}(\mathcal{D},\theta_{\text{MAP}}) + \frac{1}{2}(\theta - \theta_{\text{MAP}})(\nabla _{\theta \theta}^2 \mathcal{L}(\mathcal{D},\theta)|_{\theta = \theta_{\text{MAP}}})(\theta - \theta_{\text{MAP}})^T, 
\end{equation} where the disappearance of the first-order Taylor expansion implies that $\theta_{\text{MAP}}$ corresponds to a minimum. Therefore, $Z$ can be expressed as a Gaussian integral, for which the analytical solution is readily accessible \cite{daxberger2021laplace}:
\begin{equation}
	\label{eq:16}
	\begin{split}
	Z &= \int \text{exp}(-\mathcal{L}(\mathcal{D},\theta)) d\theta \\
	&\approx \text{exp}(-\mathcal{L}(\mathcal{D},\theta_{\text{MAP}})) \\
	&\int \text{exp}(-\frac{1}{2}(\theta - \theta_{\text{MAP}})(\nabla _{\theta \theta}^2 \mathcal{L}(\mathcal{D},\theta)|_{\theta = \theta_{\text{MAP}}})(\theta - \theta_{\text{MAP}})^T) d\theta \\
	& = \text{exp}(-\mathcal{L}(\mathcal{D},\theta_{\text{MAP}})) (2\pi)^{\frac{P}{2}}(\text{det}(\Sigma))^{\frac{1}{2}}.
    \end{split}
\end{equation}
As a result, the posterior distribution is approximated with a multivariate Gaussian distribution, i.e., $p(\theta|\mathcal{D}) \approx \mathcal{N}(\theta_{\text{MAP}}, \Sigma)$. The covariance matrix $\Sigma$ is determined by the Hessian of the posterior $\mathcal{H}$, indicated as $\Sigma = - [\mathcal{H}]^{-1}$ and $ \mathcal{H} = \nabla _{\theta \theta}^2 \mathcal{L}(\mathcal{D},\theta)|_{\theta = \theta_{\text{MAP}}}$.

Therefore, to obtain the approximate posterior using LA, it is crucial to ascertain $\theta_{\text{MAP}}$ via conventional deep learning with a proper regularizer. The only additional step involves calculating the inverse of the Hessian matrix $\mathcal{L}$ at $\theta_{\text{MAP}}$. For the log-prior regularizer $r(\theta)$ in $\mathcal{L}$, which is defined by a Gaussian prior, the Hessian is straightforward, i.e., $\beta^{-2} I$ \cite{ritter2018scalable}. Therefore, 
\begin{equation}
	\label{eq:17}
	\mathcal{H} = - N E_{(x,y)\sim\mathcal{D}}[\nabla _{\theta \theta}^2 \log p(y|x,\theta)|_{\theta = \theta_{\text{MAP}}}] -\beta^{-2} I.
\end{equation} 

\subsection{Generalized Laplace Approximation}
Here, we introduce the temperature parameter $\mathcal{T}$ to generalize the posterior distribution and obtain:
\begin{equation}
	\label{eq:18}
	p_{\mathcal{T}}(\theta|\mathcal{D}) = \frac{p(\mathcal{D}|\theta)^{\mathcal{T}}p(\theta)}{\int_{\theta'} p(\mathcal{D}|\theta')^{\mathcal{T}}p(\theta')d\theta'} = \frac{1}{Z_{\mathcal{T}}} \text{exp}(-\mathcal{L}_{\mathcal{T}}(\mathcal{D},\theta)),  
\end{equation}
\begin{equation}
	\label{eq:19}
  	\mathcal{L}_{\mathcal{T}}(\mathcal{D},\theta) = \mathcal{T}\sum_{n=1}^{N}\ell(x_n,y_n,\theta)+r(\theta),
\end{equation}

\begin{equation}
	\label{eq:20}
	Z_{\mathcal{T}} = \int_{\theta'} p(\mathcal{D}|\theta')^{\mathcal{T}}p(\theta')d\theta' = \int_{\theta'} \text{exp}(-\mathcal{L}_{\mathcal{T}}(\mathcal{D},\theta')) d\theta'.
\end{equation} 
This method allows us to derive a generalized loss function and normalization factor. By following a similar derivation as Eq. (\ref{eq:16}), we can once again analytically ascertain the value of $Z_{\mathcal{T}}$, which remains constant upon determining $\theta_{\text{MAP}}$. 

$Z_{\mathcal{T}}$ depends on $D$ and $\mathcal{T}$ but is independent of $\theta$, which follows directly.

\textbf{Proof of Invariance of $Z_{\mathcal{T}}$.}
Expand $\mathcal{L}_{\mathcal{T}}$ around the mode $\theta_{\text{MAP}}$ using a second-order Tyler expansion:
\begin{equation}
	\mathcal{L}_{\mathcal{T}}(D, \theta) \approx \mathcal{L}_{\mathcal{T}}(D, \theta_{\text{MAP}}) +\frac{1}{2}(\theta-\theta_{\text{MAP}})\mathcal{H}_{\mathcal{T}}(\theta-\theta_{\text{MAP}})^T,
\end{equation} where $\mathcal{H}_{\mathcal{T}} = \nabla^2_{\theta \theta} \mathcal{L}_{\mathcal{T}}(D, \theta)|_{\theta = \theta_{\text{MAP}}}$.

Then, 
\begin{equation}
	\begin{split}
		Z_{\mathcal{T}} & \approx \text{exp}(-\mathcal{L}_{\mathcal{T}}(D, \theta_{\text{MAP}})) \\
		& \int_{\theta'} \text{exp}(-\frac{1}{2}((\theta'-\theta_{\text{MAP}})\mathcal{H}_{\mathcal{T}}(\theta'-\theta_{\text{MAP}})^T)) d\theta'\\
		&=\text{exp}(-\mathcal{L}_{\mathcal{T}}(D, \theta_{\text{MAP}}))(2\pi)^{\frac{P}{2}}(\text{det}(\mathcal{H}^{-1}_{\mathcal{T}}))^{\frac{1}{2}}.
	\end{split}
\end{equation}
Crucially, after the quadratic expansion, the integral is a standard Gaussian integral centered at $\theta_{\text{MAP}}$ and does not depend on $\theta$ elsewhere.

This characteristic allows for generalization by straightforwardly adjusting the computation of the Hessian matrix with a temperature factor.

Given that $\mathcal{T}$ is constant, the Hessian matrix $\mathcal{H}_{\mathcal{T}}$ of $\mathcal{L}_{\mathcal{T}}(\mathcal{D},\theta)$ is evident:
\begin{equation}
	\label{eq:21}
	\mathcal{H}_{\mathcal{T}} = - \mathcal{T}N E_{(x,y)\sim\mathcal{D}}[\nabla _{\theta \theta}^2 \log p(y|x,\theta)|_{\theta = \theta_{\text{MAP}}}] -\beta^{-2} I.
\end{equation}

For uncertainty estimation at a test point $x^*$, it is conventional to employ Monte Carlo integration to approximate the intractable expression $p(y|x^*, \mathcal{D}) = \int p(y|x^*,\theta)p(\theta|\mathcal{D})d\theta$. This entails generating $S_{mc}$ samples $(\theta_{s})_{s=1}^{S_{mc}}$ from the posterior distribution of network parameters, followed by Bayesian analysis, as described in the subsequent equations:
\begin{equation}
	\label{eq:22}
	\begin{split}
		& p(y|x^*, \mathcal{D}) \approx \frac{1}{S_{mc}} \sum_{t=1}^{S_{mc}} p(y|x^*,\theta_{s}), \\
		& \theta_{s} \sim \mathcal{N}(\theta_{\text{MAP}}, -[\mathcal{H}_{\mathcal{T}}]^{-1}).
	\end{split}
\end{equation} 

Because computing or inverting the Hessian matrix is often impractical, the positive semi-definite approximations of the Hessian matrix, including Fisher information matrix and generalized Gauss-Newton, are commonly employed \cite{immer2021improving}. Additionally, dimensionality reduction techniques like Kronecker factorization \cite{ritter2018scalable} and low-rank approximation \cite{lee2020estimating} can be utilized. In \cite{chen2024sparse}, it is noted that approximating the Hessian matrix can limit the spread of uncertainty. This effect on the value of $\mathcal{T}$ will be examined in Section \ref{sec:4}.

\section{Experiments}
\label{sec:4}   
\subsection{Uncertainty Concentration Induced by Approximation} 
\label{sec:4.1}
\begin{figure}
	\centering
	\subfloat[Diag]{
		\includegraphics[width=0.23\textwidth]{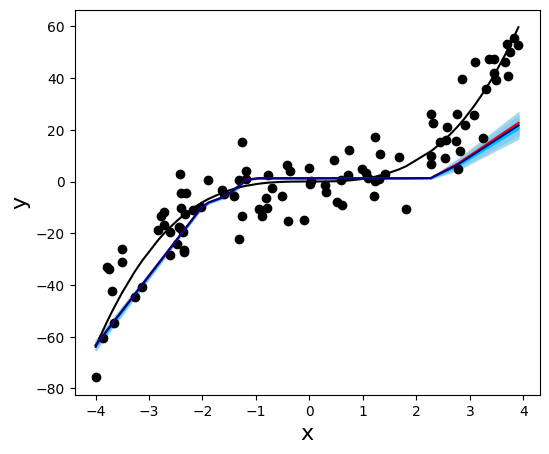}
	}
	\subfloat[KFAC]{
		\includegraphics[width=0.23\textwidth]{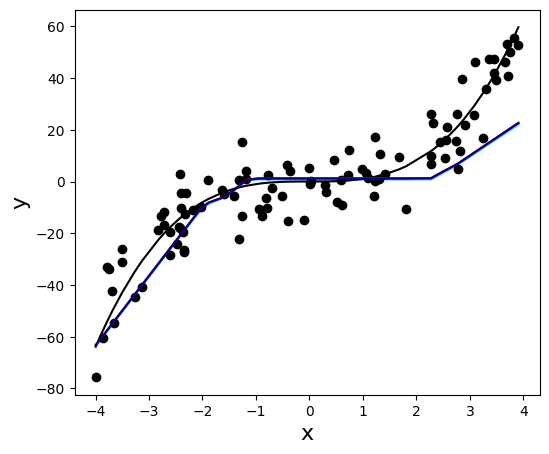}
	}
	\hspace{0.1em}
	\subfloat[EKFAC]{
		\includegraphics[width=0.23\textwidth]{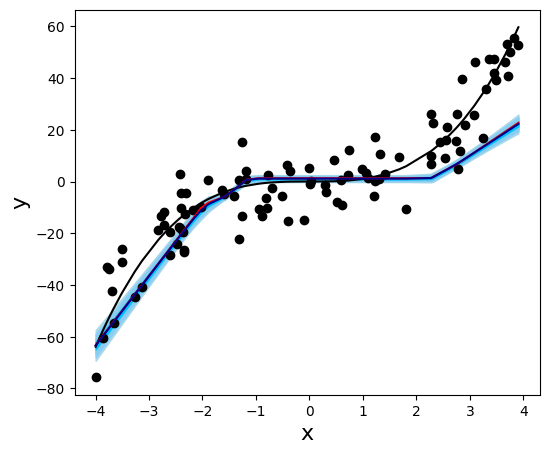}
	}
	\subfloat[B-Diag]{
		\includegraphics[width=0.23\textwidth]{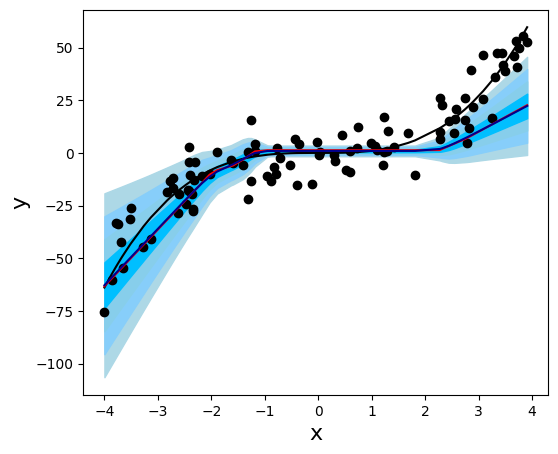}
	}
	\caption{Uncertainty visualization for a toy regression dataset under $\mathcal{T} = 1$. Black dots represent data points, and a noiseless cubic function is shown by the black line. The red line indicates the deterministic prediction, while the blue line represents the mean of the predictive distribution. Additional standard deviations are depicted by various shades of blue.}
	\label{fig:hessianApprox}
\end{figure}

First, we visualize uncertainties arising from various Hessian approximations including diagonal approximation (Diag), Kronecker-factored approximate
curvature (KFAC) \cite{ritter2018scalable}, eigenvalue-corrected Kronecker factorization (EKFAC) \cite{george2018fast}, block-diagonal approximation (B-Diag) \cite{martens2015optimizing}, on a toy regression dataset, similar to \cite{ritter2018scalable, hernandez2015probabilistic}. For detailed explanations of the Hessian approximations used in this experiment, please refer to \cite{chen2024sparse}. We generate a dataset with 100 uniformly distributed points, $x\sim U(-4,4)$, corresponding targets $y = x^3 + \epsilon$, where $\epsilon \sim N (0, 9)$. A two-layer neural network with seven units per layer is then used for model fitting. Fig. \ref{fig:hessianApprox} illustrates the uncertainty of the same neural network resulting from diagonal, Kronecker-factored, eigenvalue-corrected Kronecker-factored, and block-diagonal Laplace methods. Note that the EKFAC method approximates the Hessian matrix with higher precision compared to KFAC, as detailed in \cite{george2018fast}, Appendix A. Additionally, B-Diag offers the highest approximation precision among the mentioned methods, as it only isolates each layer within the neural network, assuming no correlation between weights across layers, a fundamental assumption of the other methods. Meanwhile, the estimated uncertainty generally follows the order: $\text{KFAC} < \text{Diag} < \text{EKFAC} < \text{B-Diag}$.  We can have a preliminary hypothesis that the approximation of Hessian matrix can lead to the concentration of uncertainty, with finer approximations resulting in greater uncertainty. We will not examine this hypothesis in this paper, but we will evaluate how this effect influences the performance of the GLA in the following section. 

\subsection{In-Distribution Experiments} 
\label{sec:4.2}
\begin{figure}
	\centering
	\subfloat[Accuracy]{
		\includegraphics[width=0.41\textwidth]{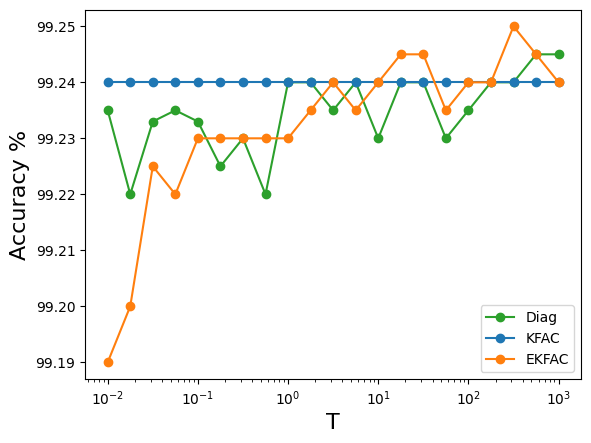}
	}
	\hfill
	\subfloat[Entropy]{
		\includegraphics[width=0.41\textwidth]{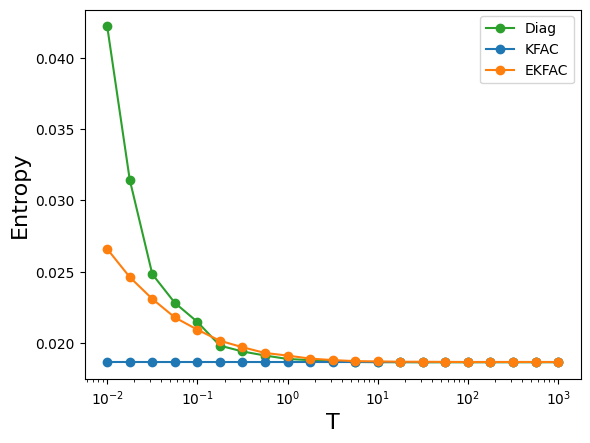}
	}
	\caption{Predictive performance for a generalized LeNet-5 model on MNIST dataset. Note that the results for both in-distribution and out-of-distribution experiments are derived using Bayesian model averaging with a sample size of 50.}
	\label{fig:coldposterior}
\end{figure}

\begin{figure}
	\centering
	\subfloat[Accuracy]{
		\includegraphics[width=0.45\textwidth]{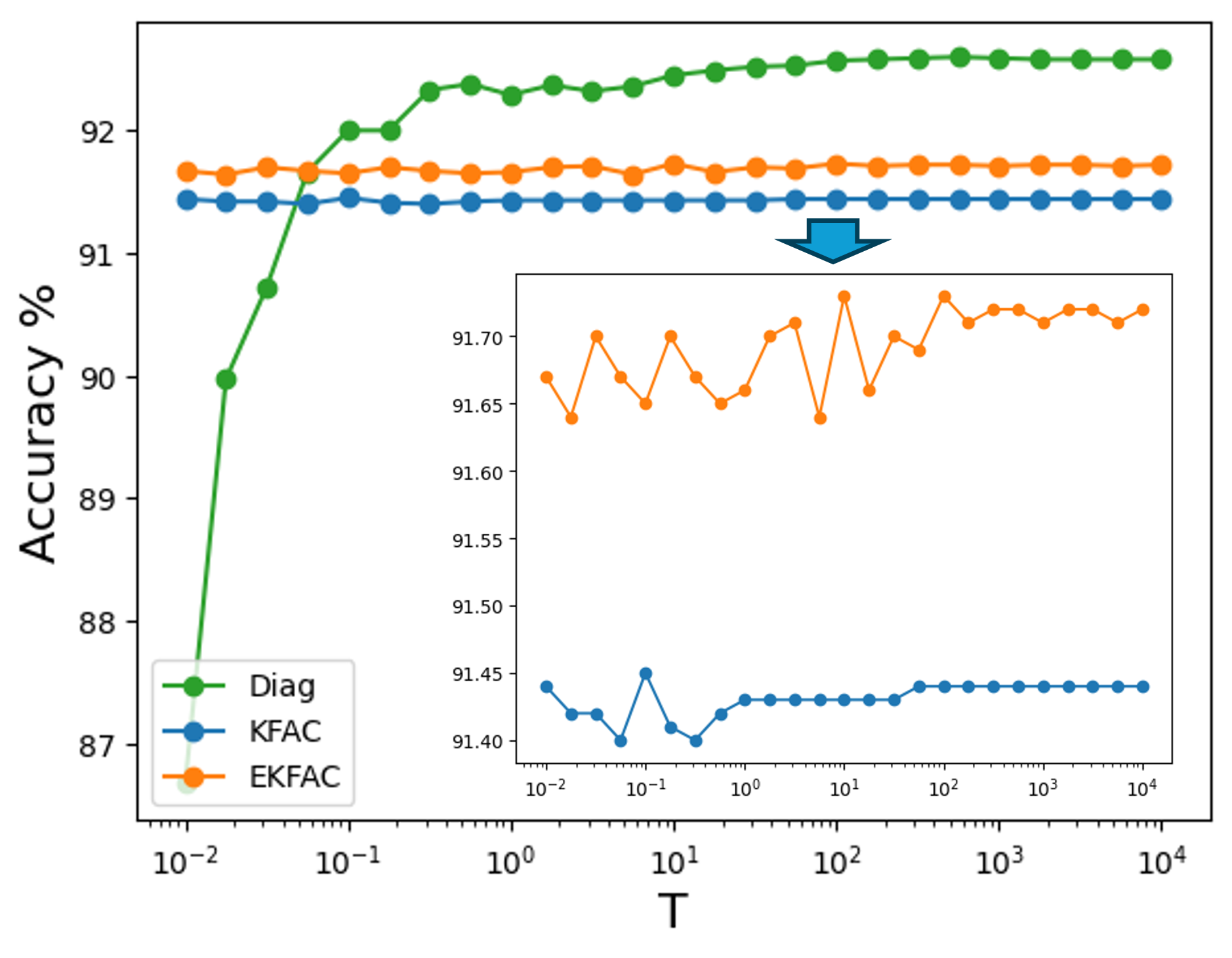}
	}
	\hfill
	\subfloat[Entropy]{
		\includegraphics[width=0.45\textwidth]{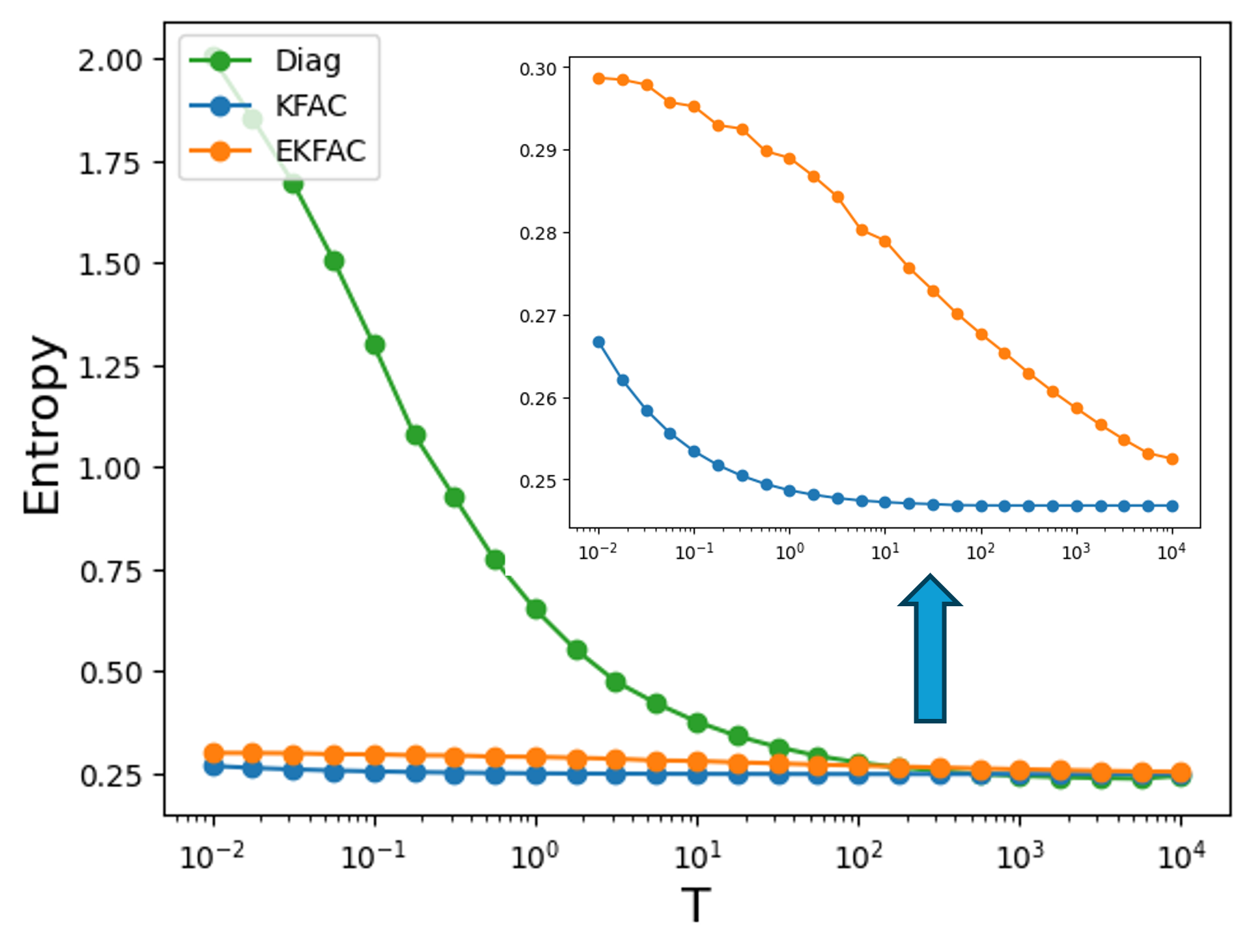}
	}
	\caption{Predictive performance for a generalized ResNet-18 model on CIFAR-10 dataset. The curves of KFAC and EKFAC are magnified.}
	\label{fig:coldposteriorres}
\end{figure}

Next, we analyze the performance of the GLA under varying $\mathcal{T}$ values and different Hessian approximations. We exclude B-Diag from this analysis, because even though B-Diag considers the covariance matrix of only one layer, in modern neural networks, its size remains large, being quadratic to the number of parameters in the layer. In this experiment, we use LeNet-5 \cite{lecun1998gradient} with ReLU activation and L2 regularization to fit the MNIST dataset, and ResNet-18 \cite{he2016deep} for the CIFAR-10 dataset \cite{krizhevsky2009learning}. Notably, a fine-tuned pre-trained model is used to align more closely with the previously discussed theory, i.e., the pre-trained model can be regarded as  $\theta_{\text{MAP}}$.

Fig. \ref{fig:coldposterior} and \ref{fig:coldposteriorres} depict the performance, measured by accuracy and entropy, of the GLA under varying $\mathcal{T}$ values and different Hessian approximations. It is observed that both accuracy and entropy improve as $\mathcal{T}$ increases for all three Hessian approximations. Specifically, the accuracy and entropy of Diag and EKFAC generally increase with $\mathcal{T}$ and converge at $\mathcal{T} > 1$, consistent with the findings in \cite{wenzel2020good}. Meanwhile, KFAC converges at $\mathcal{T}$ slightly greater than 1 in Fig. \ref{fig:coldposteriorres} and at $\mathcal{T} < 1$ in Fig. \ref{fig:coldposterior}. We attribute this to KFAC's approximation, which results in extremely large Hessian matrix values (e.g., in the LeNet-5 experiment, the Frobenius norm of the Hessian matrix of the last layer is $1.0985\times 10^9$, significantly larger than that of Diag and EKFAC). The extremely large Hessian matrix values result in two outcomes: 1) the impact of $\mathcal{T}$ is diminished, due to the huge difference in magnitude between the Hessian of the likelihood and the priori, and 2) the variance of the posterior distribution is down-scaled, causing the uncertainty concentration phenomenon discussed in Section \ref{sec:4.1}. The issue with KFAC can arise from its computation using the Kronecker product of two matrices derived from input and backpropagation gradient whereas Diag's computation relies solely on the gradient, and EKFAC involves a re-scaling. 

\subsection{Out-of-Distribution Experiments}    
\begin{figure}
	\centering
	\subfloat[Entropy of a generalized LeNet-5 model trained on MNIST evaluated on Kuzushiji-MNIST dataset.]{
		\includegraphics[width=0.45\textwidth]{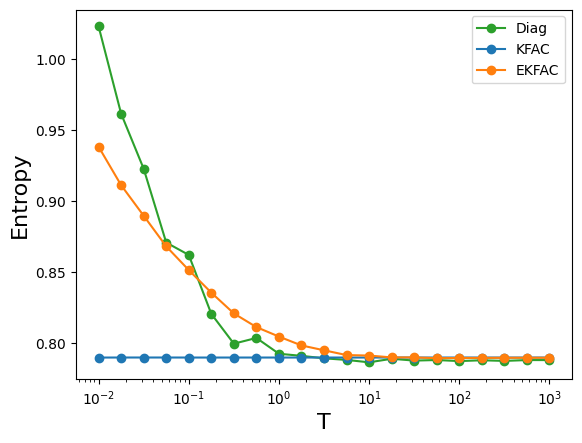}
	}
	\hfill
	\subfloat[Entropy of a generalized ResNet-18 model trained on CIFAR-10 evaluated on SVHN dataset.]{
		\includegraphics[width=0.45\textwidth]{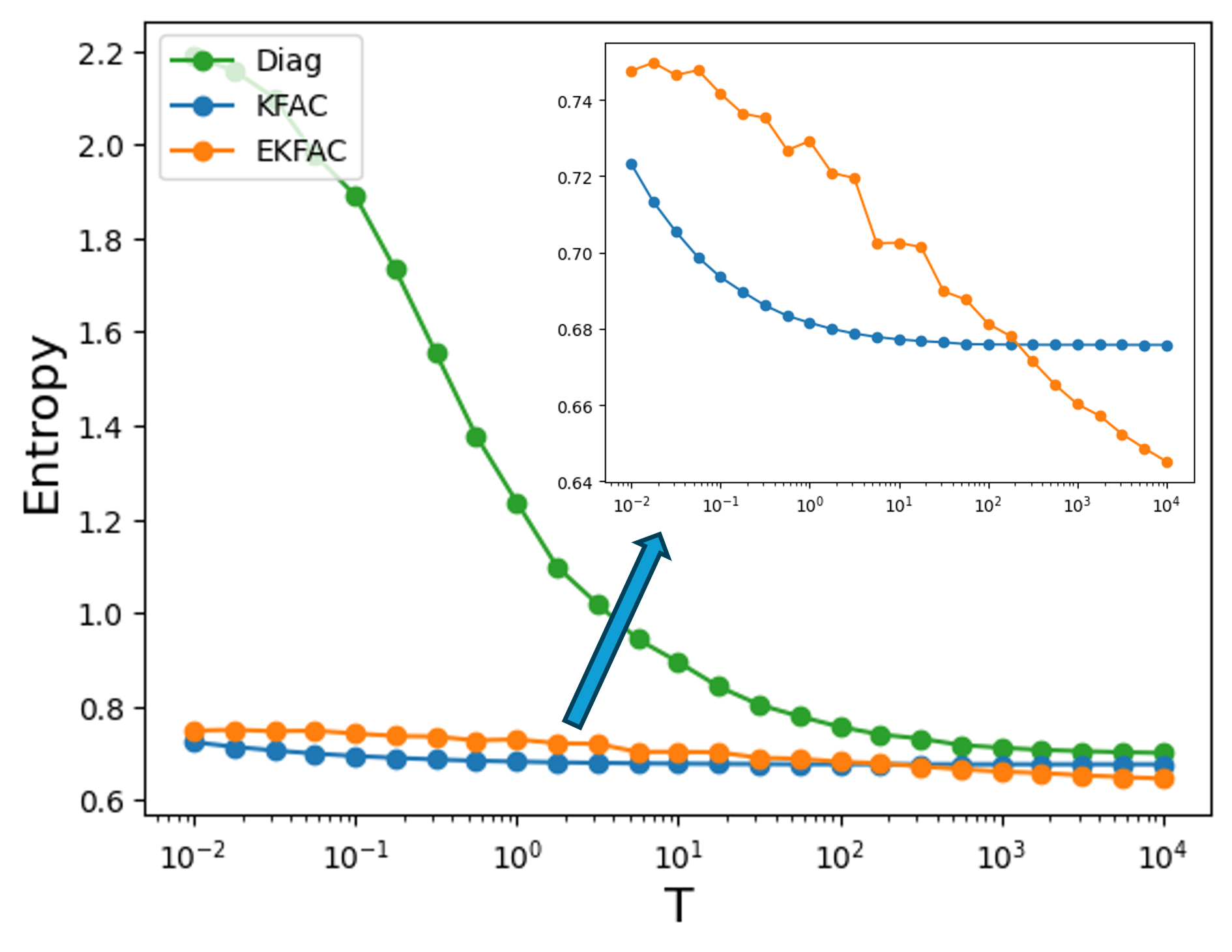}
	}
	\caption{Uncertainty estimation of generalized Laplace approximation on out-of-distribution datasets.}
	\label{fig:ood}
\end{figure}

In Section \ref{sec:4.2}, we demonstrate that generalized Bayes can enhance the performance of LA with an appropriately chosen $\mathcal{T}$. However, the optimal choice of $\mathcal{T}$ is significantly influenced by the selected Hessian approximation. In this section, we extend our investigation of GLA to out-of-distribution data. This aspect has been underexplored in existing research on generalized Bayes, due to the common assumption that Bayesian deep learning is inherently well-suited for such tasks based on its epistemic uncertainty estimation capacity.

In this study, we use the LA models trained in Section \ref{sec:4.2} to evaluate out-of-distribution datasets. Specifically, we assess LeNet-5 trained on the MNIST dataset using the Kuzushiji-MNIST test set \cite{clanuwat2018deep} and ResNet-18 trained on CIFAR-10 using the SVHN test set \cite{netzer2011reading}. However, in contrast to in-distribution experiments, where we expect lower entropy,  reflecting higher confidence in predictions on familiar data, in out-of-distribution, higher entropy is expected, indicating that the classifier produces more uniform predictions across its classes on unfamiliar data. 

Fig. \ref{fig:ood} presents the predictive uncertainties of GLA models on out-of-distribution datasets. The GLA demonstrates behavior consistent with in-distribution experiments. However, since high entropy is expected to yield uniform predictions, a small $\mathcal{T}$ is preferred for out-of-distribution data. This can be attributed to the susceptibility of Bayesian deep learning to out-of-distribution data. \cite{d2021uncertainty} propose that there exists a fundamental conflict between achieving strong generalization performance by integrating priors with high uncertainty when dealing with out-of-distribution data. According to \cite{izmailov2021dangers}, Appendix G, it is found that the temperature factor can enhance performance on out-of-distribution data arising from Gaussian noise corruption, but does not notably improve performance in cases of other types of test data corruption (note that the out-of-distribution data in this experiment is caused by domain shift).

In the context of LA, this phenomenon can be intuitively understood. The LA models the posterior distribution as a multivariate normal distribution, with the covariance matrix primarily influencing the estimation of epistemic uncertainty. This covariance matrix is derived from the inverse of the Hessian matrix of the regularized loss. The generalized covariance matrix can be expressed as follows:
\begin{equation}
	-[\mathcal{H}_{\mathcal{T}]}^{-1} = (\mathcal{T} M + \tau I)^{-1}, 
\end{equation} where $\mathcal{T}$ is a constant, $M = N E_{(x,y)\sim\mathcal{D}}[\nabla _{\theta \theta}^2 \log p(y|x,\theta)|_{\theta = \theta_{\text{MAP}}}]$ is a matrix, $\tau = \beta^{-2}$ is a constant, and $I$ is the identity matrix. By factoring out $\mathcal{T}$ from the inverse, we obtain
\begin{equation}
	\label{eq:22}
	-[\mathcal{H}_{\mathcal{T}]}^{-1} = \frac{1}{\mathcal{T}}(M + \frac{\tau}{\mathcal{T}} I)^{-1}. 
\end{equation} Eq. (\ref{eq:22}) shows that increasing $\mathcal{T}$ contracts the posterior covariance matrix, concentrating the posterior around the MAP estimate. However, in very high-dimensional settings or with ill-conditioned Hessians $\mathcal{H}$, the scaling effect of $\mathcal{T}$ can interact with the spectrum of $\mathcal{H}$ in nontrivial ways, potentially leading to numerical instability or excessively sharp posteriors.

\section{Conclusion}
This study develops a theoretical framework for how generalized Bayes addresses model misspecification and suboptimal priors. Additionally, we propose a method called GLA, which integrates the concept of generalized Bayes into LA. With an appropriate choice of $\mathcal{T}$, this method enhances in-distribution predictive performance, although the optimal $\mathcal{T}$ selection is notably impacted by the chosen Hessian approximation. Furthermore, we empirically demonstrate the vulnerability of the GLA to out-of-distribution data induced by domain shift.

An evident expansion of this research is to explore new strategies to address the susceptibility of the GLA of out-of-distribution data. For example, integrating an appropriate prior could offer a solution to this issue, such as a Gaussian process prior with a radial basis function kernel to disregard prior knowledge, or employing the data empirical covariance prior \cite{izmailov2021dangers} to prevent the posterior from aligning with the prior along corresponding parameter space directions.

\bibliographystyle{ieeetr}
\bibliography{refer.bib}

\end{document}